\makeatletter\def\graphicscache@inhibit{true}\makeatother

\documentclass[letterpaper, 10 pt, conference]{ieeeconf}  %

\IEEEoverridecommandlockouts                              %

\usepackage[utf8]{inputenc}

\usepackage[hidelinks]{hyperref}
\usepackage[style=ieee,hyperref,natbib=true,backend=bibtex,firstinits,doi=false,%
     mincitenames=1,maxcitenames=2,maxbibnames=99,sorting=none,terseinits=false,hyperref=true,defernumbers=true]{biblatex}
\bibliography{references.bib}
\renewbibmacro*{bbx:savehash}{}%
\defbibheading{bibliography}[\bibname]{\section*{References}}

\usepackage{url}
\usepackage{graphicx}
\usepackage[usenames,dvipsnames]{xcolor}
\usepackage{fixme}
\fxsetup{theme=color}

\usepackage{amsmath}
\usepackage{amssymb}
\usepackage{textcomp}

\usepackage{booktabs}
\usepackage{threeparttable}
\usepackage{multirow}

\usepackage[capitalize]{cleveref}

\usepackage{tikz}
\usetikzlibrary{arrows}
\usetikzlibrary{positioning,calc}
\usetikzlibrary{decorations.pathreplacing}
\usetikzlibrary{decorations.markings}
\usetikzlibrary{fit}
\usetikzlibrary{shapes.callouts}
\usetikzlibrary{shapes.geometric}
\usetikzlibrary{matrix}
\usepackage{setspace}

\usepackage{pgfplots}
\pgfplotsset{compat=1.9}
\usepgfplotslibrary{groupplots}
\usepgfplotslibrary{units}

\usepackage{balance}

\usepackage{blindtext}

\usepackage{ifthen}

\usepackage[export]{adjustbox}

\usepackage{graphicscache}

\title{\LARGE \bf
Fast Object Learning and Dual-arm Coordination\\ for Cluttered Stowing, Picking, and Packing
}

\author{Max Schwarz$^{*}$, Christian Lenz, Germán Martín García, Seongyong Koo, \\ Arul Selvam Periyasamy, Michael Schreiber, and Sven Behnke%
\thanks{$^{*}$All authors are with the Autonomous Intelligent Systems group of University of Bonn, Germany; {\tt schwarz@ais.uni-bonn.de}}%
}

\begin{document}

\maketitle
\thispagestyle{empty}
\pagestyle{empty}

\begin{tikzpicture}[remember picture,overlay]
\node[anchor=north west,align=left,font=\sffamily,yshift=-0.2cm] at (current page.north west) {%
  In: Proceedings of the International Conference on Robotics and Automation (ICRA) 2018
};
\node[anchor=north east, align=right,font=\sffamily,yshift=-0.2cm] at (current page.north east) {%
  DOI: \href{https://doi.org/10.1109/ICRA.2018.8461195}{10.1109/ICRA.2018.8461195}
};
\end{tikzpicture}%

\begin{abstract}

Robotic picking from cluttered bins is a demanding task, for which Amazon Robotics holds challenges.
The 2017 Amazon Robotics Challenge (ARC) required stowing items into a storage system, picking specific items, and packing them into boxes.
In this paper, we describe the entry of  team NimbRo Picking.
Our deep object perception pipeline can be quickly and efficiently adapted
to new items using a custom turntable capture system and transfer learning. It produces high-quality item segments, on which grasp poses are found.
A planning component coordinates manipulation actions between two robot
arms, minimizing execution time.
The system has been demonstrated successfully at ARC,
where our team reached second places in both the picking task and the final stow-and-pick task.
We also evaluate individual components.

\end{abstract}

\section{Introduction}

In order to successfully approach robotic bin picking, multiple research fields
ranging from computer vision to grasp planning, motion planning, and execution control need
to be tightly coupled. Especially the case of cluttered bin picking,
i.e. the picking of randomly arranged items of different types, is the focus of active research.
In order to advance the state of the art,
Amazon hold annual challenges: The Amazon Picking Challenges (APC) 2015 and 2016,
and the Amazon Robotics Challenge (ARC) 2017\footnote{\url{http://phx.corporate-ir.net/phoenix.zhtml?c=176060&p=irol-newsArticle&ID=2290376}}.

On a high level, the ARC required contestants to solve two common warehouse
tasks: The stowing of newly arrived items into a storage system (``stow task''), and the retrieval
and packing of specific items from storage into boxes (``pick task'').
In contrast to the APC~2016, the 2017 ARC allowed participants
much more leeway with regards to system design. In particular, the storage system
itself could be built by the teams.
On the other hand, the task was made more challenging by not providing all items
to the teams well before the competition, instead requiring participants to learn
new items in short time (45\,min). This forced the development
of novel object perception approaches.

\begin{figure}
 \centering
 \includegraphics[width=\linewidth,clip,trim=200 0 200 0]{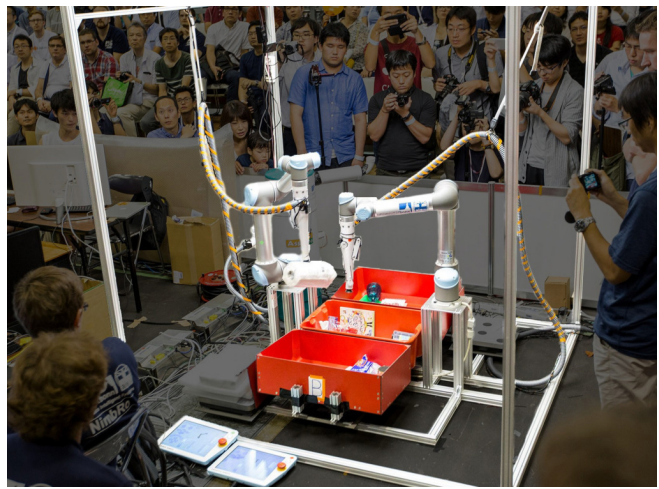}
 \caption{Our system at the Amazon Robotics Challenge 2017 in Nagoya, performing the stow phase of the final task.
 Image by Amazon Robotics.}
 \label{fig:system_at_arc}
\end{figure}

Our team NimbRo Picking developed a robotic system for the ARC 2017 (see \cref{fig:system_at_arc}). Contributions include:

\begin{itemize}
 \item A method for quickly and efficiently capturing novel items with
   minimal human involvement,
 \item a highly precise deep semantic segmentation pipeline which can be
   adapted to new items on-the-fly, and
 \item a method for online dual-arm coordination planning and execution control for complex picking or stowing tasks.
\end{itemize}

\section{Related Work}

The Amazon Picking Challenge 2016 resulted in the development of some very
interesting systems for bin picking, serving as inspiration for our system.

\Citet{Hernandez2016team} won the picking and stowing challenges.
Their system consisted of an industrial arm equipped with a hybrid
suction and pinch gripper. The team also used, like a number of other teams, a fixed camera setup
for perception of items in the tote---allowing the perception pipeline to run
while the robot is putting an item away. This motivated us to build a
fixed sensor gantry for our system.

\Citet{matsumotoend} placed second in the pick task and fourth in the
stow task. Their system directly trains a neural network to predict
item grasp poses. We initially decided against such an approach
because item grasp annotations would be expensive to obtain for
new items and we were not sure whether grasp affordances could be effectively
transferred from the known items.

Our own entry for the Amazon Picking Challenge 2016 \citep{schwarz2017nimbro, schwarz2017object}
placed second in the
stow competition and third in the pick competition.
It used a single UR10 arm, could only use suction for manipulation,
and required manual annotation of entire tote or shelf scenes for training
the object perception pipeline.

In recent years, research on semantic segmentation advanced significantly. Large datasets allow the training of increasingly
complex models (e.g. \citep{lin2016refinenet,chen2016deeplab}), but few
works focus on fast item capture and training, as required for ARC.

Dual-arm manipulation has been investigated for a long time, mostly inspired
by the human physiology. \Citet{smith2012dual} survey different approaches
and introduce the useful distinction of goal-coordinated manipulation (two
arms working towards a shared goal without direct physical interaction) and
bimanual manipulation (two arms manipulating the same item).
In this scheme, our system falls into the former category.
Since most works focus on the bimanual manipulation case \citep{bersch2011bimanual}, \citep{hebert2013dual} or consider
sequential manipulation of one item with two arms \citep{harada2012pick}, our case of online coordination of independent
manipulation in a shared workspace is underresearched. 
Other works focus on collision-free multi-robot manipulation planned offline \citep{akella2002coordinating}.
This is not sufficient in our case, since the arm trajectories and timings are not fully known in advance.

\section{Mechatronic Design}

Our system design was driven by three objectives: task completion, speed, and
simplicity (in this order). It was important to focus on task completion first,
since any time bonus would only be awarded if the task was complete.
We figured that it would be likely that
only few, if any, teams would complete the entire
task---indeed, in the finals no team completely stowed and picked all required items.

\subsection{Arms and Grippers}

Our experience from APC~2016 year told us that suction is a very powerful tool for
bin picking---we could manipulate all items using suction at APC~2016, but had 
to develop special grasp motions for specific items. Since for ARC~2017 half of the items
in the competition were unknown beforehand,
we wanted to be prepared for items requiring mechanical grasping.
To this end, we developed a hybrid gripper, similar to other
top APC 2016 teams.

To address our second design goal, speed, we developed a dual-arm
system. In particular the pick task lends itself to parallelization---three
cardboard boxes have to be filled with specific items, which can be done
mostly independently as long as multiple target items are visible, i.e. not
occluded by other items.

\begin{figure}
 \centering
\begin{tikzpicture}
    \node[anchor=south west,inner sep=0] (image) at (0,0) {\includegraphics[height=5cm]{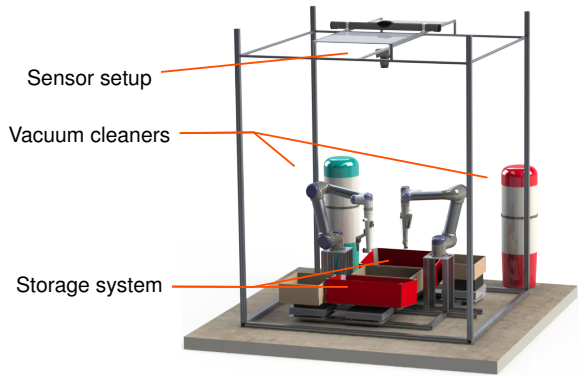}};
    \begin{scope}[
            x={(image.south east)},
            y={(image.north west)},
            font=\sffamily,
            line/.style={line width=.7pt,red!70!yellow},
            box/.style={rectangle,black},
	    every node/.style={align=center,font=\footnotesize\sffamily}
          ]

        \node[box,anchor=south] (a) at (-0.2,0.2) {Storage system};
		\draw[line] (a) -- (0.2,0.25) -- (0.43,0.25);
		\draw[line] (a) -- (0.2,0.25) -- (0.53,0.32);
        \node[box,anchor=south] (b) at (-0.2,0.61) {Vacuum cleaners};
		\draw[line] (b) -- (0.2,0.66) -- (0.3,0.57);
		\draw[line] (b) -- (0.2,0.66) -- (0.77,0.54);

        \node[box,anchor=south] (c) at (-0.2,0.75) {Sensor setup};
		\draw[line] (c) -- (0.43,0.87);

    \end{scope}
\end{tikzpicture}
\vspace*{-2ex}
 \caption{CAD model of the entire system.}
 \label{fig:system}
\end{figure}

Our robot system consists of two 6\,DoF Universal Robots UR5 arm. Each arm is equipped with
an end effector with a bendable suction finger and a 2-DoF second finger (see \cref{fig:gantry+eef}).
The high air flow needed for imperfectly sealed suction grasps is generated by two powerful vacuum cleaners, one for each arm. The vacuum can be released
through an actuated bleed valve.

\subsection{Storage System}

\begin{figure}
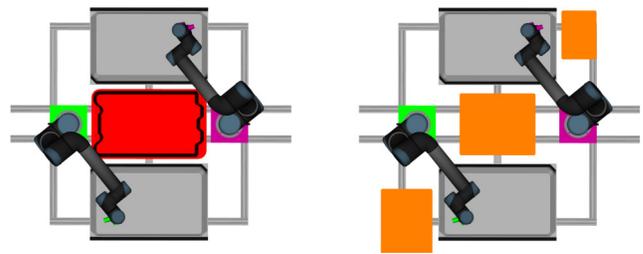

 \includegraphics[height=3.3cm,trim=0 30 0 70]{images/system_stowing_color.png}\hfill
 \includegraphics[height=3.3cm,trim=0 30 0 70]{images/system_picking_color.png}\vspace*{-1ex}
 \caption{System setup for both tasks. Storage system bins are depicted in gray. Left: Configuration with tote (red) for the stow task.
 Right: Cardboard boxes (orange) for the pick task.}
 \label{fig:setup_pick_n_stow}
\end{figure}

The Amazon-provided red tote contains the items to be stored.
It is easily accessible from the top for perception and manipulation.
Our storage system meets the maximum allowed volume and area constraints.
To match the perception and manipulation situation to the tote,
we divided the available volume into two similarly shaped bins and made then red as well.
Both parts of the storage system are reachable by both arms (see \cref{fig:setup_pick_n_stow}) and are tilted by approx. $5^\circ$ towards the center of the robot system
to increase the visibility of items located close to the inner walls of the bins. The tote (stow task) or one of the cardboard boxes (pick task) is placed between the
two storage system bins. One remaining cardboard box for the pick task is placed next to each arm and is only accessible by this arm.
An industrial scale (1\,g absolute precision) is mounted under each of the five possible pick and place locations for detecting contact and for confirming picks by checking the item weight.

\subsection{Gantry Sensors}

\begin{figure}
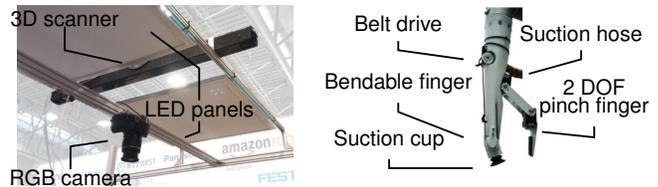

\begin{tikzpicture}
    \node[anchor=south west,inner sep=0] (image) at (0,0) {\includegraphics[height=2.4cm]{images/gantry.jpg}};
    \begin{scope}[
            x={(image.south east)},
            y={(image.north west)},
            font=\sffamily,
            line/.style={line width=.7pt,black},
            box/.style={rectangle, black},
	    every node/.style={align=center,font=\footnotesize\sffamily,inner sep=0}
          ]

        \node[box,anchor=south] (a) at (0.2,0.88) {3D scanner};
		\draw[line] (a) -- (0.2,0.75) -- (0.4,0.68);
        \node[box,anchor=south] (b) at (0.22,0) {RGB camera};
		\draw[line] (b) -- (0.22,0.22) -- (0.35,0.25);
        \node[box,anchor=south] (c) at (0.67,0.36) {LED panels};
		\draw[line] (c) -- (0.67,0.3) -- (0.62,0.27);
		\draw[line] (c) -- (0.67,0.65) -- (0.5,0.85);

    \end{scope}
\end{tikzpicture}
\hfill
\begin{tikzpicture}
    \node[anchor=south west,inner sep=0] (image) at (0,0) {\includegraphics[height=2.4cm]{images/gripper3.JPG}};
    \begin{scope}[
            x={(image.south east)},
            y={(image.north west)},
            font=\sffamily,
            line/.style={black, line width=.7pt},
            box/.style={rectangle,black},
	    every node/.style={align=center,font=\footnotesize\sffamily}
          ]

        \node[box,anchor=south] (sf) at (-0.3,0.46) {Bendable finger};
		\draw[line] (sf) -- (-0.3,0.41) -- (0.2,0.27);
        \node[box,anchor=south] (sc) at (-0.37,0.15) {Suction cup};
		\draw[line] (sc) -- (-0.37,0.1) -- (0.27,0.1);

        \node[box,anchor=south] (pf) at (1.2,0.45) {2 DOF};
        \node[box,anchor=south] (pf) at (1.2,0.33) {pinch finger};
		\draw[line] (pf) -- (1.2,0.27) -- (0.8,0.3);

        \node[box,anchor=south] (sh) at (1.1,0.75) {Suction hose};
		\draw[line] (sh) -- (1.1,0.7) -- (0.7,0.62);

		\node[box,anchor=south] (sh) at (-0.3,0.8) {Belt drive};
		\draw[line] (sh) -- (-0.3,0.75) -- (0.2,0.7);

    \end{scope}
\end{tikzpicture}
  \caption{Left: Gantry setup. Right: 3\,DoF suction gripper.}
 \label{fig:gantry+eef}
\end{figure}

For workspace perception, our robot system is equipped with a 24\,MPixel photo camera (Nikon D3400) and a 3.2\,MPixel Photoneo PhoXi\textsuperscript{\textregistered} 3D-Scanner XL (see \cref{fig:gantry+eef}). 
The 3D scanner offers sub-millimeter absolute accuracy on well-measurable surfaces\footnote{\scriptsize\url{http://photoneo.com/product-detail/phoxi-3d-scanner-xl}}.
Both sensors are mounted
on a gantry approx. 2\,m above the storage system and tote. This configuration allows for observing the entire workspace without moving the sensors.
Two LED panels provide diffuse lighting to reduce the influence of uncontrolled outside light.

\section{Object Perception}

\subsection{Item Capture \& Modeling}
\label{sec:data_capture}

\begin{figure}
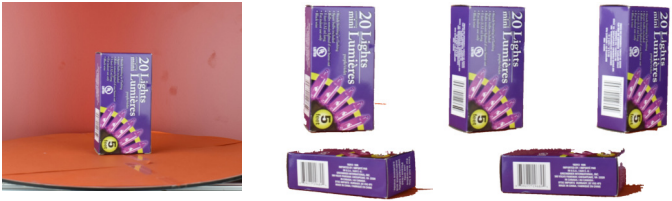

 \centering
 \adjustbox{valign=t}{\includegraphics[height=2.5cm]{images/perception/objects/light_string/0000/0000/rgb_cut.jpg}}
 \adjustbox{valign=t}{\begin{minipage}{.6\linewidth}~
 \includegraphics[height=1.8cm,clip,trim=0 60 0 70]{images/perception/objects/light_string/conv_0000.png}~
 \includegraphics[height=1.8cm,clip,trim=0 60 0 70]{images/perception/objects/light_string/conv_0001.png}~
 \includegraphics[height=1.8cm,clip,trim=0 60 0 70]{images/perception/objects/light_string/conv_0002.png}\\ 
 \includegraphics[height=1cm,clip,trim=0 0 0 100]{images/perception/objects/light_string/conv2_0002.png}~
 \includegraphics[height=1cm,clip,trim=0 0 0 100]{images/perception/objects/light_string/conv2_0003.png}
 \end{minipage}}\vspace*{-1ex}
 \caption{Turntable capture and automatic segmentation.
 Left: Input image. Right: Extracted segments in standing and lying configuration.}
 \label{fig:turntable}
\end{figure}

\newlength{\segmheight}
\setlength{\segmheight}{5.7cm}

During a competition run, our system has to quickly adapt to the provided new items.
We experimented with using only the few images provided by Amazon, but obtained
significantly better results using more images (see \cref{sec:eval:sematic_segmentation}).
The key issue is that capturing tote scenes and annotating them manually
as in our 2016 system \citep{schwarz2017nimbro} would be much too time
consuming.

Instead, we capture item views using an automated turn\-table setup (see \cref{fig:turntable}),
as used for many RGB-D object datasets~\citep{singh2014bigbird}.
The turntable is equipped with a Nikon D3400 camera and LED panels
(as on the gantry) and an Intel RealSense SR300 RGB-D sensor.
It captures twenty
views from all sides in 10\,s. A typical
item can be scanned in three different resting poses in about
a minute (including manual repositioning), resulting in sixty views.

Before starting the item capture, we also record a frame without the item.
We then use a background subtraction scheme to automatically obtain a binary
item mask. The masks are visualized and the mask generation parameters can
be quickly tuned using a graphical user interface.
The background color is exchangeable, but we did not need to do so
during the ARC 2017 competition---even objects with red parts could be reliably
extracted from red background after manual tuning of the extraction thresholds.
Note that a red background creates similar effects on transparent objects
as if they were placed in the red tote or storage system.

\subsection{Semantic Segmentation Architecture}

In our previous APC 2016 entry~\citep{schwarz2017nimbro}, we demonstrated gains
from enhancing semantic segmentation with results from object detection, which
produces less spatial detail in its bounding box outputs, but has a better notion of ``objectness''
and detects entire object instances---which helps to eliminate spurious segmentation results.
For ARC 2017, we decided to go with a pure semantic segmentation pipeline.
This decision was motivated by i) the small gain obtained by the hybrid pipeline
and ii) the fact that Amazon removed the possibility of multiple items of the same class
being in the same container, making true instance segmentation unnecessary.
In our experience, having pixel-precise segmentation instead of just bounding box-based
object detection is a big advantage for scene analysis and grasp planning.

As a basis, we reimplemented the RefineNet architecture proposed by \citet{lin2016refinenet},
which gave state-of-the-art results on the Cityscapes dataset. It uses
intermediate features from a pretrained ResNet-101 network~\citep{he2016identity},
extracted after each of the four ResNet blocks.
Since the features become more abstract,
but also reduce in resolution after each block, the feature maps are sequentially
upsampled and merged with the next-larger map, until the end result
is a both high-resolution and highly semantic feature map. The classification
output is computed using a linear layer and pixel-wise SoftMax.
For our purposes, we replaced the backbone network with the similar but newer
ResNeXt-101 network \citep{Xie2016}.

\subsection{Cluttered Scene Synthesis \& Fast Training}

\newlength{\sceneheight}
\setlength{\sceneheight}{2.5cm}

\begin{figure}
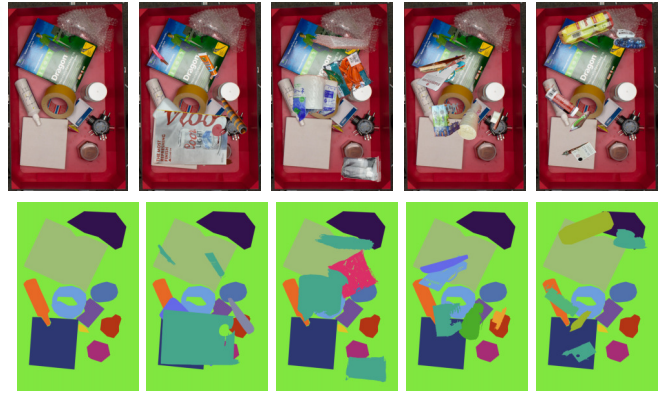

 \includegraphics[height=\sceneheight]{images/perception/scenes/ais_friday/0000/rgb.jpg}\hfill
 \includegraphics[height=\sceneheight]{images/perception/scenes/ais_friday/scenes/0/output.png}\hfill
 \includegraphics[height=\sceneheight]{images/perception/scenes/ais_friday/scenes/1/output.png}\hfill
 \includegraphics[height=\sceneheight]{images/perception/scenes/ais_friday/scenes/2/output.png}\hfill
 \includegraphics[height=\sceneheight]{images/perception/scenes/ais_friday/scenes/3/output.png}
 
 \vspace{1ex}

 \noindent\includegraphics[height=\sceneheight]{images/perception/scenes/ais_friday/orig/label_color.png}\hfill
 \includegraphics[height=\sceneheight]{images/perception/scenes/ais_friday/scenes/0/label_color.png}\hfill
 \includegraphics[height=\sceneheight]{images/perception/scenes/ais_friday/scenes/1/label_color.png}\hfill
 \includegraphics[height=\sceneheight]{images/perception/scenes/ais_friday/scenes/2/label_color.png}\hfill
 \includegraphics[height=\sceneheight]{images/perception/scenes/ais_friday/scenes/3/label_color.png}\vspace*{-1ex}
 \caption{Generated synthetic scenes. All scenes were generated with the same annotated
 background frame (left column) for easier comparison.
 Top row: RGB. Bottom row: Color-coded generated segmentation ground truth.
 }
 \label{fig:scenes}
\end{figure}

\begin{figure*}
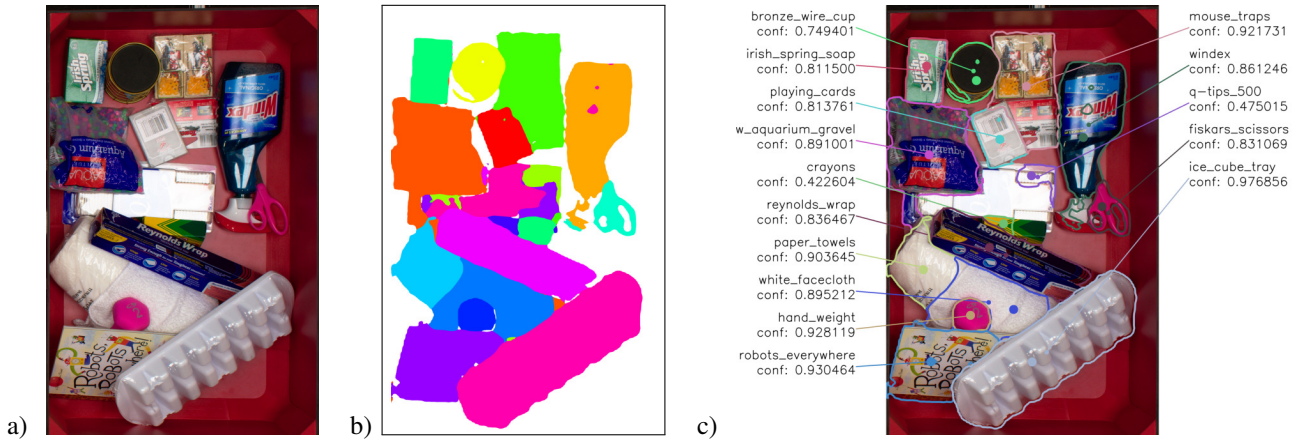

 \centering
 \setlength{\fboxsep}{0cm}
 \begin{tabular}{ccc}
  a) \includegraphics[height=\segmheight]{images/perception/2017_07_30_08_48_40/SSB/rgb.png} &
  
  b) \fbox{\includegraphics[height=\segmheight]{images/perception/2017_07_30_08_48_40/segmentation_colored.png}} &
  
  c) \includegraphics[height=\segmheight]{images/perception/2017_07_30_08_48_40/vis.png} \\
 \end{tabular}
 \caption{Object perception example from the picking phase of our finals run at ARC 2017.
 The original model trained during the run was used.
 a) RGB image captured by the Nikon camera.
 b) Segmentation output.
 c) Processed item contours with average confidences, polygon center of mass (small points), and suction spots (large points).
 Best viewed in color.
 }
 \label{fig:perception_examples}
\end{figure*}

As mentioned above, a key requirement for our system is the fast adaption to new items.
Since the amount of training images we can capture is very limited
and the item images are recorded on a turntable
without occlusions, we generate new synthetic scenes for training (see \cref{fig:scenes}).

This scene generation is done on-the-fly during training, so that we can
immediately start training and add new turntable captures as they become available.
Manually annotated dataset frames are used as background, with five new items
placed randomly on top. The new items are allowed to occlude each other to simulate
complex arrangements.
The scene generation part runs purely on CPU and is multithreaded to achieve
maximum performance. As the scene generation is faster than CNN training, we can
generate a new scene for every training iteration---ensuring that the
model does not overfit to specific arrangements.

The network training itself is distributed over $N$ GPUs.
We train on $N$ images (one image per card) and then average and synchronize
the weight gradients using the NCCL library\footnote{\url{https://github.com/NVIDIA/nccl}}.
Using one scene generation pipeline per GPU card, we can obtain 100\% GPU utilization.
During ARC 2017, the network was trained on four Titan X (Pascal) cards.

While the ResNeXt backbone network is kept fixed during training, all other RefineNet layers
and the final classification layer are trained with a constant learning rate.
Weight updates are computed using the Adam optimizer \citep{kingma2014adam}.
We pretrain the network on the set of 40 known objects, and then finetune during the competition
for the new objects.
After every epoch, the filesystem is scanned for new turntable captures
and, if required, the classification layers are adapted to a new number of classes
by extending the weight tensors with random near-zero values.

\subsection{Heuristic Grasp Selection}

Since it is infeasible to manually specify grasp positions for the
large number of items, especially for the new items in each competition run,
we built a robust grasp pose heuristic for 2D grasp proposal.
The heuristic is tuned towards suction grasps.
To avoid the dangerous item boundaries, the heuristic starts with the contour
predicted by the segmentation pipeline. As a first guess, it computes
the point with maximum distance $d_p$ to the item contour, the so-called pole of
inaccessibility~\citep{garcia2007poles}. For fast computation, we use an approximation
algorithm\footnote{\url{https://github.com/mapbox/polylabel}}.

For most lightweight items, the pole of inaccessibility suffices, which reduces the risk of missed grasps.
For heavy items, it is more important to grasp close to the center of mass.
To this end, we also check the 2D polygon center of mass and compute its distance $d_m$
to the contour. If $\frac{d_m}{d_p} > \tau$, we prefer to grasp at the center of mass.
We use a threshold $\tau\!=\!0.8$ for lightweight and $\tau\!=\!0.4$ for heavy items (weight$>$800\,g).
See \cref{fig:perception_examples} for examples.

In order to generate a 5D suction pose (rotations around the suction axis are not considered),
depth information is needed. We upsample and filter the depth map generated by the PhoXi 3D scanner by projecting
it into the camera frame and running a guided upsampling filter \citep{ferstl2013image, schwarz2017object}.
The resulting high-resolution depth map is used to estimate local surface normals.
Finally, the 5D suction pose consists of the 3D grasp contact point and the local surface normal.

For pinch grasps, the rotation around the suction axis has to be determined.
Here, we point the second finger towards the bin center, to avoid collisions.
We add Gaussian noise
on translation ($\sigma=1.5\,\textrm{cm}$) and rotation ($\sigma=60^\circ$), in order to obtain
different grasp poses for each manipulation attempt.

\subsection{The Clutter Graph}

\begin{figure}
 \centering
 \includegraphics[width=\linewidth,clip, trim=0 550 0 0,compress=false]{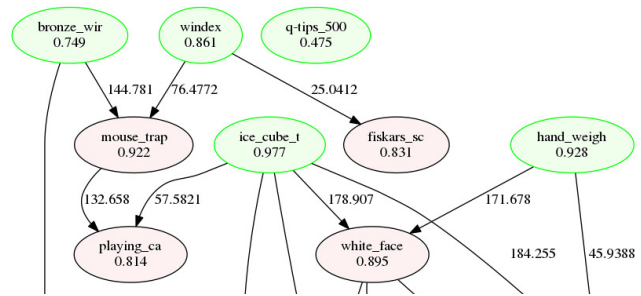}\vspace*{-1ex}
 \caption{Clutter graph for the scene in \cref{fig:perception_examples}. The bottom half is cut off,
 leaving only the items on top of the pile.
 Vertices contain the class name and detection confidence.
 Green vertices have no predecessor.
 Edges are labelled with the point count (predecessor higher than successor).
 }\vspace{-1ex}
 \label{fig:clutter_graph}
\end{figure}

For high-level planning, it is quite important to estimate which items are currently
graspable and which are occluded by other items that would need to be removed to get access to the item of interest.
For this reason, we generate a directed graph that we call {\em clutter graph}.
All perceived items are vertices in this graph, with an edge from $A$ to $B$ indicating
that $A$ is occluding $B$. See \cref{fig:clutter_graph} for an example.

The graph is initially generated by examining the item contours.
Along the contour, we check the upsampled depth map for points on the outer side which are higher
than the corresponding points on the inner side. These points are counted and an edge
is inserted into the graph, directed from the occluding item to the item under consideration.
The point count (as evidence for the occlusion) is attached to the edge.

After simplifying cycles of length two (edges and back edges) by reducing them to one
edge with the difference in point counts, the graph may still contain cycles,
which would block certain items from ever being removed.
We resolve this situation by deleting the set of edges with minimum point count sum (i.e. minimum
evidence) that makes the graph acyclic.
This is called the minimum feedback arc set and is NP-hard~\citep{karp1972reducibility}, but for our small graphs we can quickly compute a brute-force solution.
This method both reliably removes cycles caused by measuring errors and forces an acyclic solution
in cases where the occlusion is actually cyclic---allowing the system to try and extract an item
from the cycle, since no other action is possible.
The result is a directed acyclic graph containing the occlusion information.

\subsection{Object Pose Estimation}

\begin{figure}
 \centering
 \includegraphics[width=1\linewidth,clip,trim=0 20 0 13]{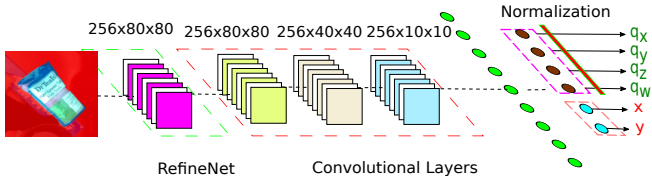}\vspace*{-1ex}
 \caption{Pose estimation network architecture.}
 \label{fig:pose_net}
\end{figure}

During preparation for the ARC 2017, we anticipated more difficult items which
would be graspable only at very specific grasp poses.
To make this possible, we developed a 6D pose estimation module, which would allow specifying
grasps relative to an item frame.
Our method does not compute a fused 3D item model, which can be difficult to obtain
for transparent objects. Instead, we train an additional CNN on predicting
the pose from individual views.

The architecture of the pose estimation network is shown in \cref{fig:pose_net}.
It predicts the 3D orientation of the item relative to the camera frame in the form
of an unit quaternion. A second branch predicts the 2D pixel location of the item
coordinate frame origin.
The network consists of the RefineNet backbone as in the semantic segmentation network,
followed by three convolution layers and two fully connected layers. %
For $M$ item classes, the network predicts $6M$ values---one quaternion and translation offset per item class.
In this way, the predictor is conditioned on the object class, which is inferred by the segmentation network.
During training, the object segment captured on the turn\-table is placed on top of a randomly cropped storage system scene.
Furthermore, the background is shifted towards red to emphasize the item currently under consideration (see \cref{fig:pose_net}).
The output of the pose estimation network is projected to a full 6D pose using the depth map.
\section{Dual-Arm Motion Generation}

\subsection{Parametrized Motion Primitives}

The UR5 arms and the endeffectors are controlled with parametrized motion primitives. A motion is defined by a set of keyframes which 
specify the kinematic group(s) manipulated by this motion. Each keyframe either defines an endeffector pose in Cartesian space or the joint state of the
kinematic group(s). 
The keyframes are either manually designed beforehand or generated and adapted to perception results at runtime.
This motion generation has been used on other robot systems in our group before (see \cite{schwarz2017nimbro} and \cite{schwarz2016drc}).

\subsection{Inverse Kinematics}

\begin{figure}
 \centering
 \includegraphics[height=.4\linewidth,clip,trim=0 50 0 0]{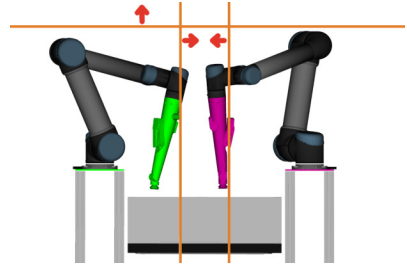}
 \caption{Cost function planes for the IK solver. The planes affect the wrist of the robot.
 The vertical plane keeps the endeffector vertical, as long as the horizontal planes are not
 active (purple robot). The horizontal planes keep the wrist away from the robot base to prevent self-collisions.}
 \label{fig:ik}
\end{figure}

For keyframes defined in Cartesian space, we use a selectively damped least square (SDLS) solver, 
as in \citep{schwarz2017nimbro}. %
Since the arm including the suction finger has seven DoF, we can optimize secondary objectives in the null-space
of the Jacobian matrix.
In our case, we want to keep the wrist as high as possible and thus keep the endeffector roughly vertical in order to
reduce the horizontal space needed while manipulating.
Hence, we define a horizontal plane above the robot and use the squared distance from the wrist to the plane as cost function.
In the stow task, two additional vertical planes are added (see \cref{fig:ik}) to prevent the wrist getting too close to the manipulator base. For further details, we refer to \citep{schwarz2017nimbro}.

\subsection{High-Level Planning for Picking}

The high-level planner for the pick task triggers the perception pipeline, processes the segmentation results and assigns manipulation tasks to the arms.
The perception pipeline is started for a particular bin whenever no possible tasks are left and the bin is not occluded by an arm.
Item detections are sorted according to a per-item fail counter, the number of items occluding the target item, and the perception confidence.
The two best ranked target items are marked as possible tasks for this bin.
If no target items are detected or the fail counter for the best items is too large, new tasks moving non-target item out 
of the way are generated.

\begin{figure}
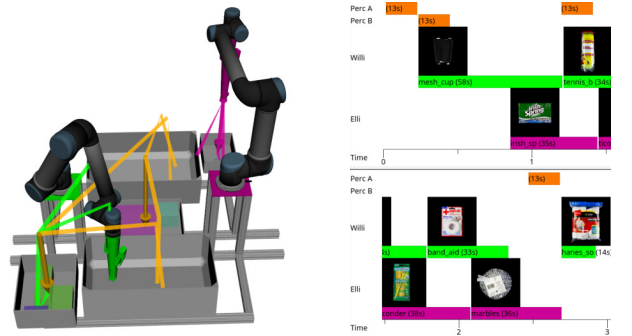
\centering
 \includegraphics[width=.45\linewidth,clip,trim=30 13 30 10]{images/pick_planner2.png}\hspace*{3ex}
 \includegraphics[width=.4\linewidth,compress=false]{images/control_gui2a.png}\vspace*{-1ex}
 \caption{Planning for the pick task. Left: Visualization of manipulation tasks.
 Chosen tasks are marked in green and purple.
 Right: Timeline of actions including perception time and arm motions.
 }
 \label{fig:pick_planner}
\end{figure}

Whenever an arm is free, we assign the best marked task considering collision avoidance with the other arm. A task consists of a set of waypoints of endeffector
poses starting with the current arm pose, grasp pose, place pose, home pose of the arm and some intermediate waypoints (see \cref{fig:pick_planner}).

Since we assume the last link of the arm to be always vertical, we only consider the 2D endeffector pose for collision checking.
Hence, all waypoints are projected into 2D.
Next, we compute the shortest Euclidean distance for each line segment defined by two
consecutive waypoints of one task to all line segments of the other task.
If the minimum of all these distances is larger than a threshold, the tasks can be
executed in parallel.
Since the number of possible tasks is limited, we can test all possible task combinations as long as an arm is free.
If multiple collision-free tasks exist, we prefer tasks which can only executed by the free arm (i.e. the place location is in one of the corner boxes).
We delete reached waypoints from current tasks to allow the second arm to start on new tasks as soon as possible.

\subsection{Placement Planning}

Since the space inside the cardboard boxes is limited, our system finds optimal placement poses inside the boxes.
The placement planner uses bounding box dimensions provided by Amazon for each item.
It considers three disjoint sets of items per box: Already placed items $\mathcal{A}$, currently possible task items $\mathcal{B}$,
and items which will be picked later $\mathcal{C}$. The planner finds a brute-force solution
in the form of a 3D stacking of the item bounding boxes, under the constraint that items
from $\mathcal{A}$ have a fixed position and items from $\mathcal{B}$ have to be placed before items from $\mathcal{C}$.
The bounding boxes may be rotated by $90^\circ$ around the vertical axis.
The solution with minimum total stacking height is then used to determine the target poses for each item from $\mathcal{B}$.

Objects of oblong shape are always placed such that the height dimension is the smallest dimension.
This may necessitate a rotating motion during placement, since the items are always grasped roughly from above.
If required, we place an additional constraint on the grasp pose which ensures that the items are grasped
in a way that allows the later rotation using our single DoF on the suction finger.

\subsection{High-Level Planning for Stowing}

In the stow task, the single pick location (the tote) limits the possibility to parallelize the manipulation work,
since precise weight change measurements require sequential picking actions in the tote.
To enable parallel work, we assign each bin of the storage system to one arm as its associated stow location.
This at least allows us to place an item with the first arm while grasping the next item with the second.

Since we have to stow all of the given items, we start with objects where we are confident that they are lying on top.
Thus, the item detections are first sorted by the confidence reported by the perception pipeline. Next, the best $\frac{3}{4}$ detections are sorted by the total number of items lying on top (from the clutter graph) and, finally, the best half of these are considered as possible tasks.

Since manipulation is performed open-loop after perception, we allow only two manipulation attempts before the
scene is measured again. We try to find a pair of items containing one of the two best detection results, respecting a minimum distance between the two items to prevent the first manipulation attempt affecting the second item.
If such a pair exists, we assign the items randomly to the arms, otherwise
we stow only the item with higher confidence.

In contrast to the pick task, no collision avoidance at the high-level planning is needed since each arm has its dedicated workspace and access to the tote is granted sequentially.

\subsection{Grasp Execution}
In both tasks (pick and stow) after each perception run, a predefined per-object probability decides which grasp type (suction or pinch) should be performed.
Our 2016 system suffered from grasping wrongly identified items during the
stowing task---a failure case which easily leads to incorrect internal states
of the high-level control, cascading the failure and creating even more problems.
Due to this experience, we wanted to eliminate false positives for ARC 2017
by double-checking perception and manipulation using a second modality.
To this end, after grasping and lifting an item, the item weight is measured with the scale mounted below the container
and compared with the expected item weight.
If the weight difference is under 5\,g or 10\% of the item weight, we accept the item and place it.
Otherwise, we drop it again and increment the fail counter.
This check also protects the system from accidentally grasping more than one item.

\section{Evaluation}

We evaluated our work on a system level at the Amazon Robotics Challenge 2017, which was held at RoboCup in Nagoya, Japan.
We augment this evaluation with separate experiments for the object perception pipeline
and the dual-arm planner.

\subsection{Amazon Robotics Challenge 2017}

\begin{table}
\caption{Timings and success rates from ARC 2017}
\centering\footnotesize\vspace*{-1ex}
\begin{threeparttable}
  \begin{tabular}{cl|rrr|rrr}
  \toprule
   & & \multicolumn{3}{|c|}{Individual Challenges} & \multicolumn{3}{c}{Final Challenge} \\
   \cmidrule (lr) {3-5} \cmidrule (lr) {6-8}
   & & \# & Time [s] & Stddev & \# & Time [s] & Stddev \\
  \midrule
  
  \parbox[t]{2mm}{\multirow{6}{*}{\rotatebox[origin=c]{90}{Stow}}}
  
   & Vision &  & & & 19 & 11.1 & 0.0 \\
   & Stows & \multicolumn{3}{|c|}{not comparable} & 14 & 29.8 & 5.4 \\
   & Fails & & & & 12 & 14.0 & 6.9\\
   \cmidrule (lr) {2-8}
   & Sum & & & & 45 & \multicolumn{2}{c}{13:17\,min} \\
   & Runtime & & & & 45 & \multicolumn{2}{c}{10:33\,min} \\
  
  \midrule
  \parbox[t]{2mm}{\multirow{6}{*}{\rotatebox[origin=c]{90}{Pick}}}
  & Vision & 13 & 12.0 & 0.9 & 32 & 13.1 & 1.3 \\
  & Picks & 10 & 38.3 & 7.6  & 8  & 39.1 & 10.0 \\
  & Moves & 4 & 34.5 & 9.0   & 10 & 30.3 & 3.0 \\
  & Fails & 5 & 20.4 & 3.2   & 22 & 28.9 & 10.9\\
  \cmidrule (lr) {2-8}
  & Sum & 32 & \multicolumn{2}{c|}{12:59\,min}    & 72 & \multicolumn{2}{c}{27:52\,min} \\
  & Runtime & 32 & \multicolumn{2}{c|}{8:56\,min} & 72 & \multicolumn{2}{c}{19:22\,min} \\ 
  \bottomrule
  \end{tabular}
\end{threeparttable}
\label{tab:times}
\end{table}

At the ARC 2017, our team had four chances to demonstrate the abilities of our system.
In our practice slot, we successfully attempted the pick task
and obtained a score of 150 points, the maximum of all practice scores.

Unfortunately, the evening before our official stow run we experienced a short in the
power supply wiring, damaging our control computer and a few microcontrollers beyond
repair. The necessary replacement and reconfiguration work did not leave us any time for full system tests before our
stow run. Consequently, due to a series
of operator mistakes caused by the new configuration, our system operated with wrong
item weights during the stow task and discarded nearly all grasped items.

We were able to fix these remaining issues for the pick task, where our team
scored 245 points, which led to a second place in the pick competition, behind
Team Nanyang with 257 points. The third placed team achieved 160 points.

Our system also performed very well in the final task, which combined the
stow and pick tasks sequentially. In the stow phase, we were able to stow
fourteen out of sixteen items. The remaining two items could not be picked
because a bug resulted in an unfortunate gripper orientation,
which was prevented from execution by collision checks.
In the picking phase, we picked eight out of ten target items.
One of the target items had not been stowed in the stow phase, so it
was impossible to pick. The other one was a cup made out of thin and sparse wires,
making it both very difficult to perceive and to grasp.
The system succeeded once in grasping it, but discarded it due to an
imprecise weight measurement. We scored 235 points in the final task, which placed
us second behind the winning Team ACRV with 272 points and in front of Team Nanyang (225 points). 

\Cref{tab:times} shows a summary of the successes and failures per task and
recorded times for perception and manipulation actions.
Generally, having two arms for manipulation lowered the overall runtime and
allowed for more manipulation attempts in a given time.
Overall, our system performed very well and showcased all of its abilities at the ARC 2017.

\subsection{Semantic Segmentation}
\label{sec:eval:sematic_segmentation}
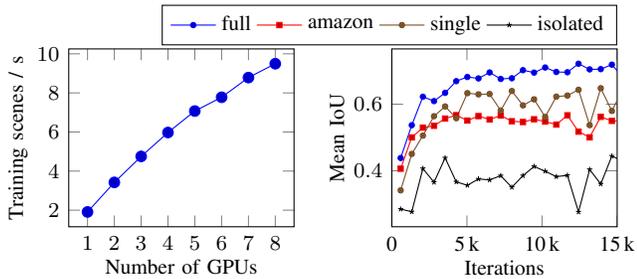
\begin{figure}
 \centering
 \begin{tikzpicture}
  \begin{groupplot}[group style={group size=2 by 1, horizontal sep=1.3cm}, every axis label/.append style={inner sep=0cm}]
   
   \nextgroupplot[font=\footnotesize, xlabel={Number of GPUs}, ylabel={Training scenes / s}, xtick distance=1, width=.53\linewidth]
  
   \addplot table [x=ngpu, y expr={144 / \thisrow{time}}] {data/gpu_benchmark.csv};

  \nextgroupplot[font=\footnotesize, xlabel={Iterations}, ylabel={Mean IoU}, width=.53\linewidth, mark size=1pt, xmin=0, xmax=15,
       xtick={0,5,10,15},xticklabels={0, 5\,k, 10\,k, 15\,k},
       legend columns=4, legend style={at={(1.0,1.02)},anchor=south east}
    ]
   \addplot+ table [x expr={4 * \thisrow{iter} / 1000}, y=informed] {data/evaluation_full.log};
   \addplot+ table [x expr={4 * \thisrow{iter} / 1000}, y=informed] {data/evaluation_amazon_only.log};
   \addplot+ table [x expr={4 * \thisrow{iter} / 1000}, y=informed] {data/evaluation_single_new_object.log};
   \addplot+ table [x expr={4 * \thisrow{iter} / 1000}, y=informed] {data/evaluation_isolated.log};

   \legend{full, amazon, single, isolated}
  \end{groupplot}
 \end{tikzpicture}%
 \caption{Segmentation experiments.
 Left: Training image throughput depending on the number of GPUs.
 Right: Test set IoU during training.
 }
 \label{fig:gpu_scaling}
\end{figure}

\begin{table}
 \centering\footnotesize
 \begin{threeparttable}
  \caption{Ablation experiments for scene synthesis}
  \label{tab:ablation}
  \begin{tabular}{lcccc}
    \toprule
    Variant          & full     & amazon   & single   & isolated \\
    \midrule
    Turntable images\tnote{1} & $\checkmark$ &          & $\checkmark$ & $\checkmark$ \\
    Rendered objects\tnote{2} & 5        & 5        & 1        & 1 \\
    Complex background\tnote{3} & $\checkmark$ & $\checkmark$ & $\checkmark$ & \\
    \midrule
    Mean IoU         & \textbf{0.720}    & 0.571    & 0.642    & 0.375 \\
    \bottomrule
  \end{tabular}
  \begin{tablenotes}
   \item [1] Otherwise the scene generation only uses Amazon images.
   \item [2] Number of rendered objects per synthetic scene. Note that one object
     means that this object is never occluded.
   \item [3] Whether complex images (full totes / storage systems) or empty
     totes are used as background for synthetic scenes.
  \end{tablenotes}
 \end{threeparttable}
\end{table}

After the ARC, we annotated the collected images during our final run
with ground truth segments to be able to quantitatively judge segmentation performance.
We then recreated the segmentation training run from our final.

To investigate the scalability of our training pipeline, we ran 10 training epochs
(with one epoch defined by 140 background images) on a varying number of Nvidia Titan Xp
cards. \Cref{fig:gpu_scaling} shows that our pipeline scales nicely to up to eight
GPUs (and possibly more).

\Cref{fig:gpu_scaling} also shows a typical test result curve recorded during training.
We measure the intersection over union (IoU) separately for each class and then average over
the classes present in the test set.
One can see that after 5,000 to 10,000 iterations the curve saturates. Using four GPUs,
as during the ARC, this occurs after approx. 15--30\,min. Note that during
a real training run, the system starts training with the images provided by Amazon
and turntable captures are added over a period of 20\,min, extending the needed time.

We performed an ablation study to determine the usefulness of individual
scene synthesis steps (see \cref{tab:ablation}). Training on scenes with objects
rendered from our own turntable images strongly outperforms using only the
Amazon-provided object images. This may be due to both insufficient number of
views (up to six in the Amazon data vs. 40--60 in ours) and our red turntable
background, yielding realistic transparency response for objects in the red tote
or storage system.
Creating occlusions on the rendered objects is important.
Finally, training on isolated scenes with only one object in an empty tote
yields poor performance, maybe indicating that our background images
with complex arrangements help regularizing the segmentation.

\subsection{Pose Estimation}

\begin{table}
\centering\footnotesize
\begin{threeparttable}
  \caption{Pose Estimation Errors}\label{tab:pose_results}
  \begin{tabular}{ccccccc}
  \toprule
   & \multicolumn{2}{c}{Translation [pix]} & \multicolumn{2}{c}{Angular [$^\circ$]}   \\
   \cmidrule (lr) {2-3} \cmidrule (lr) {4-5}
   & train & val                     & train & val \\
  \midrule
  Salts     & 2.28 & 3.32 & 1.80 & 3.19 \\
  Paper    & 2.41 & 3.79 & 1.68 & 3.09 \\
  Windex   & 2.25 & 3.41 & 1.86 & 2.78 \\
  \midrule
  Average         & 2.31 & 3.51 & 1.78 & 3.02 \\
  \bottomrule
  \end{tabular}
\end{threeparttable}
\end{table}

During the ARC 2017, pose estimation was not necessary. Our grasp heuristic was able to find good suction
or pinch grasps on all of the encountered items.
Nevertheless, we evaluated the pose estimation network by training it on three different items.
\Cref{tab:pose_results} shows quantitative results of these experiments.
Our pose estimator is able to predict the translation of the item origin within a few pixels
and the orientation within a few degrees.

\subsection{Dual-Arm Experiments}

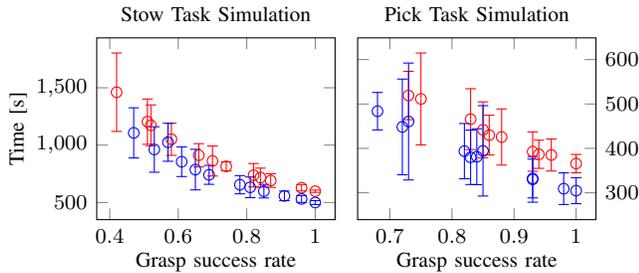
\begin{figure}
 \centering

\begin{tikzpicture}[every node/.append style={font=\footnotesize}]
\begin{groupplot}[group style={group size=2 by 1, horizontal sep=0.3cm }, height=4cm, every axis label/.append style={inner sep=0pt}, title style={inner sep=0pt}]
  \nextgroupplot[title={\footnotesize Stow Task Simulation}, width=.55\linewidth, ylabel={Time [s]}, xlabel={Grasp success rate}]
    \addplot [color=red, only marks, mark=o,]
 plot [error bars/.cd, y dir = both, y explicit]
 table[x =x, y =y, y error =ey]{data/stow_1arm.csv};\label{plots:1arm}
    \addplot [color=blue, only marks, mark=o,]
 plot [error bars/.cd, y dir = both, y explicit]
 table[x =x, y =y, y error =ey]{data/stow_2arms.csv};\label{plots:2arms}
 \coordinate (top) at (rel axis cs:0,1);%
  \nextgroupplot[title={\footnotesize Pick Task Simulation}, width=.55\linewidth, xlabel={Grasp success rate}, yticklabel pos=right]
    \addplot [color=red, only marks, mark=o,]
 plot [error bars/.cd, y dir = both, y explicit]
 table[x =x, y =y, y error =ey]{data/pick_1arm.csv};
    \addplot [color=blue, only marks, mark=o,]
 plot [error bars/.cd, y dir = both, y explicit]
 table[x =x, y =y, y error =ey]{data/pick_2arms.csv};
\coordinate (bot) at (rel axis cs:1,0);%
\end{groupplot}

\end{tikzpicture}\vspace*{-1ex}
\caption{Averaged run time with standard deviation (10 runs each) in simulation for stow and pick task with one (red) and two arms (blue) used.}
\label{fig:dual_arm_simulation}
\end{figure}

We also investigated the speedup of our system achieved by using a second arm. For both tasks (pick and stow) several full 
runs were performed in simulation.

The perception pipeline was not simulated; instead the planner was supplied with the item poses after
a certain time---11\,s for stow and 13\,s for pick, since this was the average perception time needed in the ARC final.
Object poses were generated by uniformly sampling positions inside the storage bin and the tote with fixed orientation.

We averaged the time needed for solving the task with different grasp success probability values over ten runs each. For the one arm pick task experiments, the unreachable box was symmetrically placed next to the used arm. 

\Cref{fig:dual_arm_simulation} shows the results. If only one arm is used, the system needs on average 1.2 to 1.3 times longer to complete the task. The large configuration space of grasp success rate, item locations, requested order, and order of detection
result in a high standard deviation. Nevertheless, the trend is clearly observable.

\section{Lessons Learned \& Conclusion}

The ARC 2017 also allowed us to evaluate our fundamental design choices.
Our strong focus on the object perception pipeline and efficient execution of the tasks, as opposed to more complicated
mechanical solutions, was very successful.
We also learned that even such dynamic tasks requiring fast adaption to new items are within reach of current
mainstream deep learning approaches, if one can parallelize the training and makes proper use of pretraining.

In retrospect, we could have reduced the execution time further by optimizing our storage system layout.
The dual-arm speed-up from factor 1.3 to 1.2 is slightly disappointing and is mostly limited due to resource conflicts,
e.g. both arms wanting to place in the central box. In our design, we minimized the arms' common workspace
while ensuring that storage bins and tote could be reached by both arms. However, a
different placement of boxes or more global planning could
potentially alleviate the conflicts.

As always with robotics competitions, proper full-scale testing is important, both for the system as well as the
opera\-tors. On the operator side, we made mistakes during our stow slot. On the system side, we noticed precision problems
with our scales quite late in the competition, which might have cost us the first place in the finals.

Overall, we demonstrated a successful solution for the ARC 2017.
Our object perception pipeline is able to be quickly adapted to new items, to produce precise item contours, infer grasp poses, and predict 6D item poses.
We demonstrated how to quickly plan and coordinate two arms operating in the same workspace.
Our very good results at the ARC 2017 and our quantitative experiments show the effectiveness of our approach.

\section*{Acknowledgment}

\noindent{\footnotesize This research has been supported by grant BE 2556/12-1 (ALROMA) of German Research Foundation (DFG). 
We gratefully acknowledge travel support provided by Amazon Robotics and the PhoXi\textsuperscript{\textregistered} 3D-Scanner XL, which was provided by Photoneo free of charge.}

\printbibliography

\end{document}